\documentclass{article}

\usepackage[dblblindworkshop, final]{neurips_2025}

\usepackage[utf8]{inputenc} 
\usepackage[T1]{fontenc}    
\usepackage{hyperref}       
\usepackage{url}            
\usepackage{booktabs}       
\usepackage{amsfonts}       
\usepackage{nicefrac}       
\usepackage{microtype}      
\usepackage{xcolor}         
\usepackage{graphicx} 
\usepackage{subfig}

\title{Towards Scalable Meta-Learning of near-optimal Interpretable Models via Synthetic Model Generations}
\workshoptitle{Generative AI in Finance}

\author{
  Kyaw Hpone Myint,
  Zhe Wu,
  Alexandre Day,
  Giri Iyengar \\
  Capital One \\
  \texttt{\{kyaw.hponemyint, zhe.wu, alexandre.day, giridharan.iyengar\}@capitalone.com}
}

\begin{document}

\maketitle

\begin{abstract}


Decision trees are widely used in high-stakes fields like finance and healthcare due to their interpretability. This work introduces an efficient, scalable method for generating synthetic pre-training data to enable meta-learning of decision trees. Our approach samples near-optimal decision trees synthetically, creating large-scale, realistic datasets. Using the MetaTree transformer architecture, we demonstrate that this method achieves performance comparable to pre-training on real-world data or with computationally expensive optimal decision trees. This strategy significantly reduces computational costs, enhances data generation flexibility, and paves the way for scalable and efficient meta-learning of interpretable decision tree models.
\end{abstract}

\section{Introduction}
\label{intro}
Transformer-based architectures have shown remarkable adaptability across domains like natural language processing, vision, and multi-modal tasks [\cite{NIPS2017_3f5ee243, dosovitskiy2021an, 10123038}]. A key application is in planning tasks, where models "learn to learn" by fitting priors over large problem sets and performing in-context predictions [\cite{hollmann2023tabpfn, lehnert2024beyond}]. These capabilities are particularly useful for decision-making tasks requiring a balance between exploration and exploitation.

This work focuses on meta-learning interpretable models, specifically near-optimal decision trees. Decision tree optimization is computationally challenging due to its reliance on heuristics and stochastic methods to navigate the NP-hard search space [\cite{lin2020generalized}]. While traditional tree-based methods like Gradient Boosting dominate tabular data classification, their lack of interpretability and the opaque nature of deep learning models limit their adoption in high-stakes fields like healthcare and finance [\cite{Rudin:2019tl, Amann:2020tq}]. To address these challenges, we propose leveraging Structural Causal Models (SCMs) to generate synthetic datasets for pre-training. Our approach enables scalable meta-learning of decision trees by constructing diverse datasets and training transformer-based models to achieve performance comparable to state-of-the-art methods.

%

\section{Method}
\label{method}

\subsection{Meta-learning for near optimal decision trees}
\label{metalearning4dt}

\begin{figure}[h!] 
    \centering 
    \includegraphics[width=1.0\textwidth]{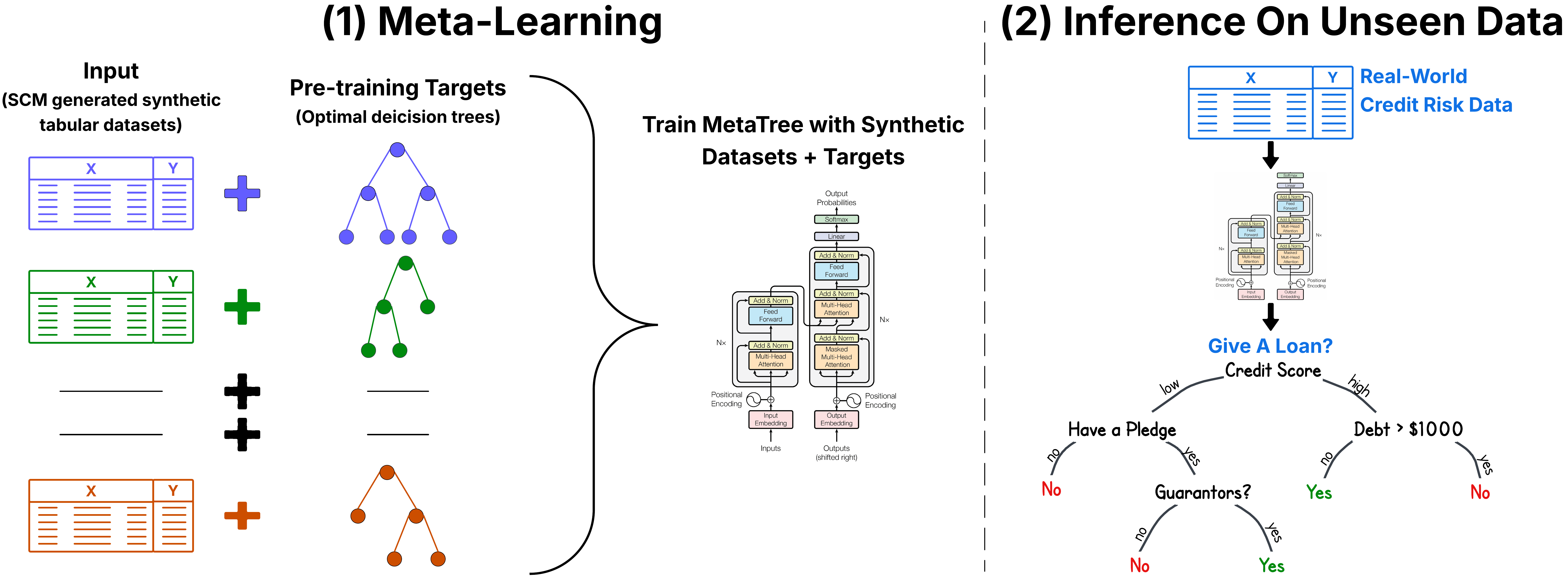} 
    \caption{Meta-learning workflow for generating look-ahead trees using synthetic data. The workflow can be divided into two parts: (1) meta-learning step where labeled synthetic datasets are fed into MetaTree along with the optimal decision trees for each dataset as the training targets, and (2) inference step where the pre-trained model is used to predict the look-ahead trees on an unseen, real-world datasets} 
    \label{fig:metalearning4trees}
\end{figure}

Fig. \ref{fig:metalearning4trees} illustrates a high-level overview of the meta-learning approach using the synthetic data generation pipeline we have developed.
This meta-learning workflow can be divided into two main steps:

\begin{enumerate}
    \item meta-learning (or pre-training) step where labeled synthetic datasets are fed into MetaTree along with the optimal decision trees for each dataset as the training targets, Fig. \ref{fig:metalearning4trees}(1)
    \item inference step where the pre-trained model is used to predict the near-optimal, look-ahead trees on an unseen, real-world datasets, Fig. \ref{fig:metalearning4trees}(2).
\end{enumerate}



During the meta-learning step, the MetaTree model is pre-trained on synthetic datasets generated using our Structural Causal Model (SCM) workflow. These datasets include features and corresponding optimal decision trees as training targets. The SCM ensures causal relationships between features and labels, while our pipeline filters out low-quality datasets using class imbalance and accuracy thresholds. This guarantees that the pre-training data is well-suited for decision tree construction. The MetaTree model learns from these synthetic datasets to predict near-optimal decision trees efficiently.

In the inference step, the pre-trained MetaTree model is applied to unseen, real-world datasets. Given a labeled dataset, the model predicts a near-optimal decision tree tailored to the data. For example, in a credit risk dataset, the model generates a decision tree to predict outcomes like "Give A Loan?" based on input features. This workflow enables scalable and interpretable decision tree generation without relying on computationally expensive optimal tree solvers or scarce real-world data.

\subsection{Structural Causal Model (SCM) based synthetic data generation workflow}
\label{scm_for_syn_data}

\begin{figure}[h!] 
    \centering 
    \includegraphics[width=1.0\textwidth]{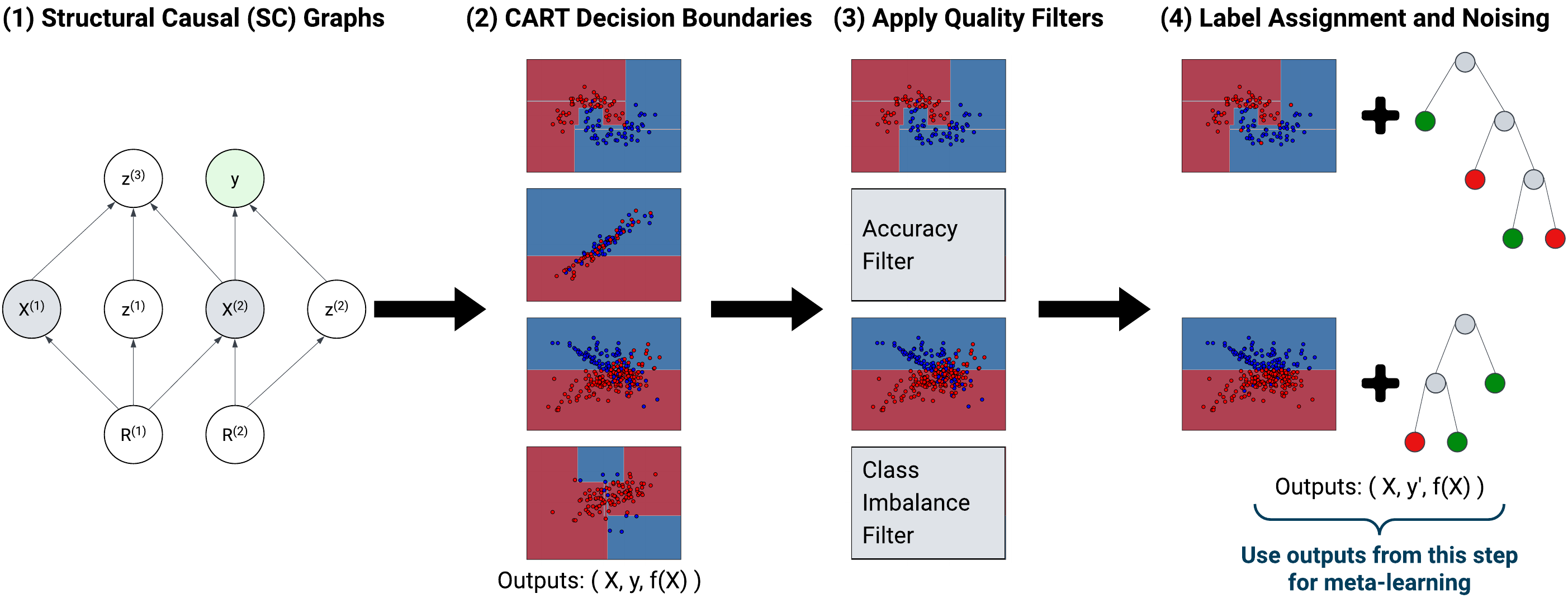} 
    \caption{Workflow for pre-training with synthetic data} 
    \label{fig:syn_data_gen_workflow}
\end{figure}

Although decision trees are widely used in high-stakes financial applications like credit scoring due to their interpretability, their construction often relies on suboptimal, greedy algorithms such as CART, which make locally optimal splits by minimizing Gini impurity or entropy. These methods lack a look-ahead mechanism, often resulting in trees that are not globally optimal. While finding an optimal decision tree is a key goal in interpretable machine learning, it is an NP-hard problem. Recent advancements like GOSDT can find certifiably optimal solutions, but their computational cost limits them to shallow trees, typically with depths of two or three. This makes generating training targets for pre-training transformer models like MetaTree prohibitively expensive, especially given the need for high-quality datasets tailored to decision tree generation.

To address this, we propose a constructive method that generates synthetic data and corresponding near-optimal trees simultaneously. Fig. \ref{fig:syn_data_gen_workflow} outlines our four-step pipeline. First, synthetic features and target labels are sampled from a Structural Causal Model (SCM) [\cite{hollmann2023tabpfn}], ensuring causal relationships between features and labels. This produces datasets $(X, y)$ with true labels.

In the second step, CART trees $f(X)$ are generated using the synthetic datasets, establishing a baseline for decision tree performance. Since not all synthetic datasets are suitable for tree generation, we apply quality filters in step three to remove datasets with issues like severe class imbalance or poor separability (e.g., Fig. \ref{fig:syn_data_gen_workflow}(3)). For instance, datasets with over 90\% of samples in one class or non-linearly separable data are discarded.

Finally, in step four, we create synthetic datasets aligned with the decision boundaries of the CART trees. This involves relabeling the original labels $(y)$ with the CART tree predictions and introducing 5\% label noise to ensure generalizability. The resulting datasets $(X, y')$ are intrinsically aligned with the trees, enabling scalable pre-training for MetaTree and faster iteration on model complexity and architecture.

\subsection{Quality filters for improving the pre-training data}
\label{qual_filters}

To enhance raw synthetic data for decision tree construction, data-quality filters were introduced to address class imbalance and ensure model accuracy. A class imbalance filter was implemented to prevent trivial decision stumps by ensuring no majority class exceeds 75\% of the samples, using a normalized class imbalance metric defined by the formulas $I = K \sum_{i=1}^K \left( \frac{n_i}{N} - \frac{1}{K} \right)^2$ which is then normalized as $I_\mathrm{normalized} = \frac{ I_\mathrm{raw} }{ I_\mathrm{worst\ case} }$. Additionally, an accuracy filter was applied, retaining only datasets where a CART tree could achieve over 70\% accuracy. These measures together select suitable datasets for generating effective pre-training targets for meta-learning. More details can be found at \ref{details_qual_filters}.

\section{Experiments and Results}
\label{experiments_and_results}

This section can be divided into four main parts. First, we will explore the scalability and diversity of the SCM based synthetic data generation workflow we have developed in Sec. \ref{scm_scalability_and_quality} and \ref{scm_diversity}.
Then, in Sec. \ref{metatree_scaling_laws}, we will explore the scaling law for MetaTree using the synthetic data we have generated.
Finally, in Sec. \ref{accuracy_benchmark} we compare the inference accuracy of a MetaTree model trained on our synthetic data against several benchmarks: the original MetaTree model (trained on hand-curated real-world data), CART, and GOSDT.

\subsection{Scalability of SCM Data Generation Pipeline}
\label{scm_scalability_and_quality}
\label{scm_scalability}

\begin{figure}[h!] 
    \centering
    \includegraphics[width=0.9\textwidth]{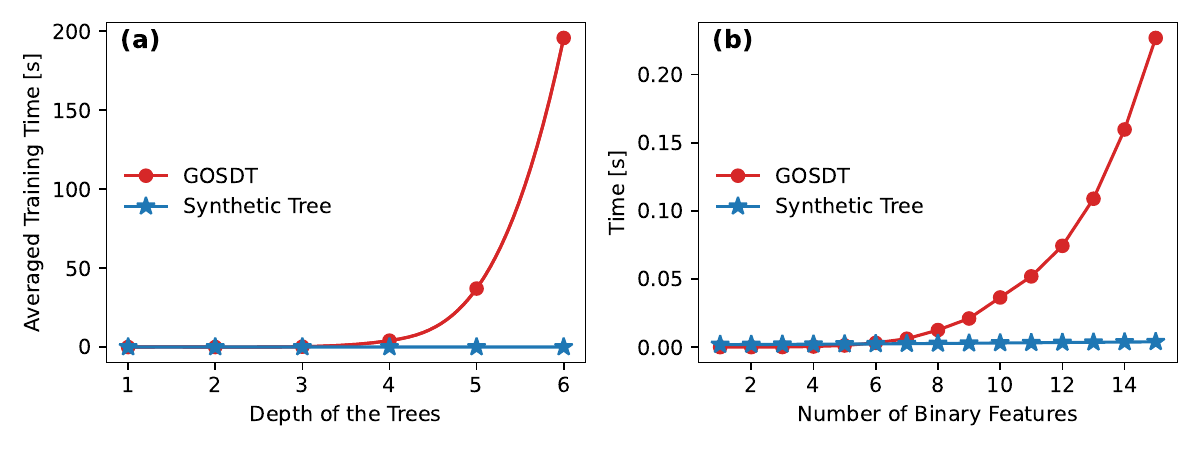}
    \caption{Time complexity as a function of: (a) the depth of the tree, and (b) the number of binary features for generating pre-training targets}
    \label{fig:time_complexity_vs_tree_deth_bfeatures}
\end{figure}

The computational complexity of generating optimal decision trees grows combinatorially with the number of features in the dataset [\cite{lin2020generalized}]. We compare the time complexity of the state-of-the-art GOSDT solver with our proposed synthetic data generation method. As outlined in Sec. \ref{scm_for_syn_data}, our method uses a CART-based algorithm with label reassignment and noising to efficiently create near-optimal decision trees. Fig. \ref{fig:time_complexity_vs_tree_deth_bfeatures}(a) shows that GOSDT's training time increases drastically with tree depth, reaching nearly 200 seconds for a depth of 6, while our method remains constant at under 1 second regardless of depth.

Similarly, Fig. \ref{fig:time_complexity_vs_tree_deth_bfeatures}(b) illustrates the impact of the number of binary features on training time. GOSDT exhibits exponential growth due to the combinatorial nature of feature selection, while our method maintains consistently low computation time. This scalability advantage stems from avoiding exhaustive searches over the feature space, making our approach significantly faster for generating near-optimal decision tree targets. These results demonstrate that our method circumvents the combinatorial complexity inherent in optimal tree algorithms, offering a practical and efficient alternative for pre-training decision tree models.

\subsection{MetaTree Scaling Laws with Synthetic Data}
\label{metatree_scaling_laws}

One common bottleneck in pretraining foundation models is the high-quality data curation process which can be time consuming, cost prohibitive, and very labor intensive.
Our proposed synthetic data generation process circumvents this problem by modeling underlying causal relationships between the input features and target labels allowing us to generate high quality synthetic datasets that can be readily used for pre-training.
In this section, we investigate the scaling properties of the MetaTree model using the structural causal model (SCM) based synthetic data generation pipeline we have developed. More details can be found in Sec. \ref{sec:scaling_law_details}.


\subsection{Average Inference Accuracy Benchmarks}
\label{accuracy_benchmark}

\begin{table}[h!] 
    \centering 
    \caption{Average accuracy and standard error of the mean (in parentheses)} 
    \label{tab:avg_accuracy} 

    \begin{tabular}{lcccc} 
        \toprule 
        \# of trees & MetaTree Original & MetaTree Synthetic Data & CART & GOSDT \\
        \midrule 
        1 & 0.6508 (0.0068) & 0.6443 (0.0070)  & 0.6502 (0.0072) & 0.6524 (0.0072) \\
        5 & 0.6783 (0.0063) & 0.6755 (0.0064) & 0.6814 (0.0063) & 0.6670 (0.0069) \\
        10 & 0.6769 (0.0061) & 0.6707 (0.0063) & 0.6806 (0.0061) & 0.6646 (0.0069) \\
        30 & 0.7047 (0.0059) & 0.6956 (0.0061) & 0.7053 (0.0060) & 0.6943 (0.0066) \\
        \bottomrule 
    \end{tabular}
\end{table}

Table~\ref{tab:avg_accuracy} summarizes the average classification accuracy and standard errors for four models: the original MetaTree, MetaTree trained on synthetic data, CART, and GOSDT. Benchmarked on 91 datasets from the original MetaTree paper [\cite{zhuang2024learning}], the models were evaluated with varying ensemble tree counts (1, 5, 10, 30) over 10 trials per dataset. Results show that the MetaTree trained on synthetic data performs comparably to the original MetaTree. For instance, with 30 trees, the synthetic MetaTree achieved 0.6956±0.0061 accuracy, closely matching the original MetaTree's 0.7047±0.0059. This validates the effectiveness of the synthetic data generation process in capturing key statistical and causal properties of real-world data without requiring costly data curation. Among benchmarks, CART achieved the highest accuracy (0.7053±0.0060 with 30 trees), while GOSDT performed well with a single tree but lagged at higher complexities, being outperformed as the number of trees increased.

\section{Conclusions}
\label{conclusions}


We introduced a scalable framework for generating synthetic pre-training data that enables the meta-learning of near-optimal decision trees. By leveraging Structural Causal Models (SCMs), our approach produces diverse, realistic datasets while circumventing the computational bottlenecks of traditional methods. A novel label reassignment and noising scheme obviates the need for expensive optimal tree solvers (e.g., GOSDT), and our experiments confirm the pipeline's constant time complexity regardless of tree depth or feature count. Complemented by targeted quality filters to ensure dataset diversity and balance, our method removes the critical dependency on high-quality real-world data, thereby enabling rapid architectural iteration and highly scalable model training.

%
%
%
%
%
%
%
%


\bibliography{references}
\bibliographystyle{neurips_2025}

\appendix


\section{Related work}
\label{Related Work}
Transformer-based meta-learning has emerged as a powerful paradigm for "learning to learn" algorithms, where models are pre-trained on a vast number of problems to generate solutions for new, unseen tasks at inference time through in-context predictions.
This approach has been successfully applied to various planning and optimization problems [\cite{zhuang2024learning, lehnert2024abetterplanningtransformers, garg2023transformerslearnincontextcase, cao2025largelanguagemodelsplanning}], demonstrating the ability to produce high-performance solutions with significant speed-ups.

A key inspiration for our work is the TabPFN paper [\cite{hollmann2023tabpfn}], which demonstrated that real-world tabular data can be effectively modeled using priors from Structural Causal Models (SCMs) for meta-learning.
While powerful, TabPFN models are opaque black-boxes, which limits their use in high-stakes domains requiring interpretability.
However, this work motivates us to leverage SCM for generating high quality, scalable tabular for pre-training purposes.

Recently, \cite{zhuang2024learning} explored meta-learning interpretable models directly with their MetaTree architecture.
Their approach, however, faced limitations in obtaining transparent and scalable training data, and relied on hand-curated datasets for pre-training the meta-learning model, which can be cost prohibiting for training larger and more predictive models.
Moreover, the pre-training targets were generated from hand-curated real-world datasets using either standard heuristic algorithms (CART) or optimal decision tree solvers (GOSDT), and this can be a cost-prohibitive way to obtain the near-optimal decision tree for pre-training purposes.

Our work takes inspiration from these prior works, and we combined SCM based synthetic data generation process to design a scalable workflow that can generate high quality tree datasets along with their corresponding near-optimal trees that can later be used in the meta-learning process.
Our proposed meta-learning workflow allows for the development of computationally efficient, near-optimal decision trees for solving high-stake financial problems.

\section{Technical Appendices and Supplementary Material}
\subsection{SCM Data Generation setup}

We generated the Structural Causal Model (SCM) data using the TabPFN v1 codebase and its provided example scripts. As TabPFN v1 is computationally lightweight, each data generation process was executed on a compute node with 16 CPU cores and 16 GB of RAM. To accelerate data generation, we ran multiple processes in parallel, each initialized with a unique random seed.

\subsection{Model Pretraining setup}

We pre-trained the model, with randomly initialized weights, on a dataset of 20 million examples. This dataset is notably twice the size of the 10 million examples used in the original MetaTree paper. Training was performed on a single server node equipped with eight NVIDIA A100 GPUs on the GenAI ultra cluster, completing 100,000 epochs in approximately 32 hours.

\subsection{Benchmark Setup}

Our benchmark setup utilizes the 91 holdout datasets from the MetaTree paper [\cite{zhuang2024learning}]. We sampled each dataset $N$ times to generate a total of $91 \times N$ unique data splits. Each split consists of a training set with $M$ samples and a test set with $Y$ samples.

For each split, all compared models (MetaTree, Decision Tree, and GOSDT) were trained on the identical training set and evaluated on the corresponding test set. To ensure a fair comparison, hyperparameters such as maximum tree depth and the number of trees were same across all models within a given evaluation.

\subsection{More details on MetaTree scaling law with synthetic data}
\label{sec:scaling_law_details}

\begin{figure}[h!] 
    \centering
    \includegraphics[width=0.8\textwidth]{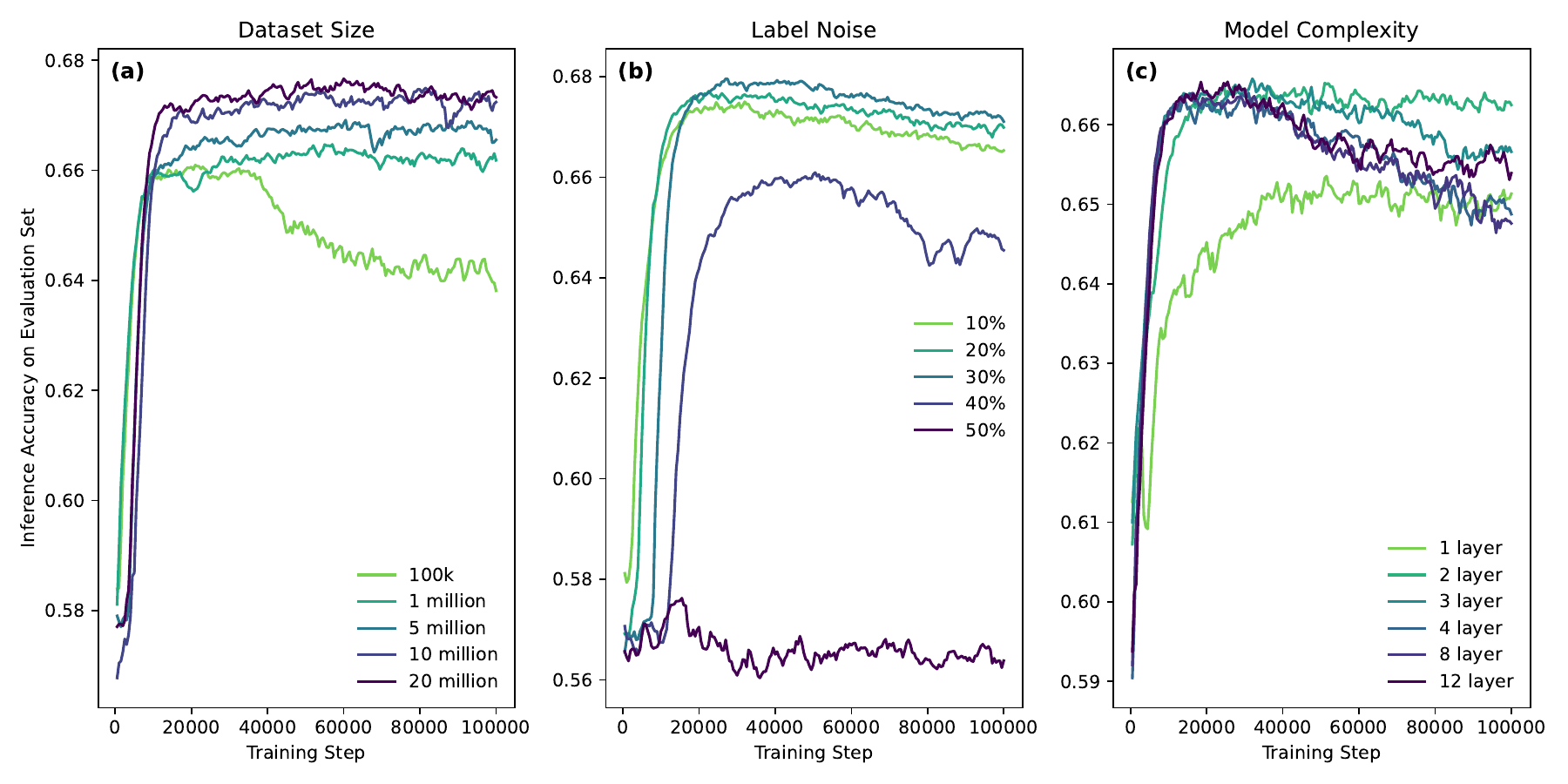}
    \caption{MetaTree scaling laws as a function of: (a) dataset size, (b) label noise, and (c) model complexity}
    \label{fig:scaling_law_all}
\end{figure}

\subsubsection{The impact of training data size on inference accuracy}
\label{data_size_on_inference_accuracy}

First, we examined the relationship between inference accuracy and the number of data points used during pre-training.
As illustrated in Fig. \ref{fig:scaling_law_all}(a), we observe a clear and positive correlation between the size of the pre-training dataset and the model's final inference accuracy on a held-out evaluation set.

The models were pre-trained on datasets of varying sizes, ranging from 100,000 to 20 million data points.
The results demonstrate a consistent trend: as the dataset size increases, the peak inference accuracy achieved by the model also increases.
The model trained on 20 million data points achieves the highest accuracy, approximately 0.675, followed in descending order by the models trained on 10 million, 5 million, and 1 million data points.
Notably, the model trained on the smallest dataset of 100,000 points not only reaches a lower peak accuracy but also exhibits instability, with its performance degrading after approximately 40,000 training steps, suggesting overfitting.
In contrast, models trained on larger datasets (5 million and above) maintain a stable and high level of performance throughout the training process.

\subsubsection{The impact of label noise on inference accuracy}
\label{label_noise_on_inference_accuracy}

Fig. \ref{fig:scaling_law_all}(b) shows the inference accuracy on an evaluation set as a function of training steps for models trained with varying levels of label noise. Five different noise levels were tested: 10\%, 20\%, 30\%, 40\%, and 50\%. Models trained with lower noise levels of 10\%, 20\%, 30\% demonstrate performance increases with increasing label noise, rapidly reaching a peak accuracy of approximately 0.68 within the first 25,000 steps.
However, as we increase the noise level further, we get diminishing return in performance.
For example, as the noise level increases from 30\% to 40\%, the learning process is slower, and the maximum achieved accuracy is lower.
The model with a noise level of 40\% peaks at an accuracy of about 0.66 after around 40,000 steps.
Moreover, the model trained with the highest noise level of 50\% fails to learn effectively, with its accuracy remaining low and fluctuating around 0.56 throughout the 100,000 training steps.
Our findings deomonstrate that a moderate level of label noise enhances the model's learning process.
However, once the noise reaches a saturation threshold, it begins to degrade inference performance, as anticipated.

\subsubsection{The impact of model complexity on inference accuracy}
\label{model_complexity_on_inference_accuracy}

As according to the power law relationship [\cite{kaplan2020scalinglawsneurallanguage}], the model performance increases predictably as a function of the model size, dataset size and the amount of compute used in training the model. 
In this section, we explore the effect of model complexity on the final performance of the model. To investigate this trade-off, we evaluated the test performance of our MetaTree model as a function of its depth, measured by the number of attention layers. The models, ranging from 1 to 12 layers, were all pre-trained on an identical dataset of 100 thousand examples to ensure a controlled comparison.

The results, presented in Fig. \ref{fig:scaling_law_all}(c), reveal a distinct "sweet spot" for model complexity for the 100k training set. The 1-layer model exhibits the lowest performance, struggling to converge to the accuracy levels of its deeper counterparts, which suggests its capacity is insufficient to capture the underlying patterns in the data. A significant performance leap is observed when increasing the depth to 2 and 3 layers. The 2-layer model, in particular, achieves the highest and most stable inference accuracy on the evaluation set, sustaining a peak performance of approximately 0.665 throughout the latter half of training.

Conversely, we observe a point of diminishing returns for models with 4 or more layers. While these deeper models (4, 8, and 12 layers) initially learn quickly, they fail to surpass the accuracy of the 2-layer model and instead show a slight but noticeable degradation in performance over extended training on the 100k-sample dataset. This suggests that for this data volume, excessive complexity introduces instabilities that hinder generalization. This finding aligns with established Chinchilla scaling law from Deepmind's 2022 paper [\cite{hoffmann2022trainingcomputeoptimallargelanguage}], which posits that model and dataset size should scale in tandem. To effectively train our 12-layer architecture—which has 6 times the depth of the optimal 2-layer model—a proportionally larger dataset would be required.

This precise characterization of the performance-complexity curve is a salient advantage of our synthetic data generation framework. By utilizing a Structural Causal Model (SCM) to generate a massive and consistent training dataset, we eliminate data variability as a confounding factor. This creates a stable testbed where the effects of architectural changes can be isolated and reliably measured. This capability to automate the generation of high-quality, causally-grounded data allows for rapid, data-driven optimization of model architecture, ensuring we can train models that are not only powerful but also optimally efficient.

\begin{figure}[h!] 
    \centering
    \includegraphics[width=0.8\textwidth]{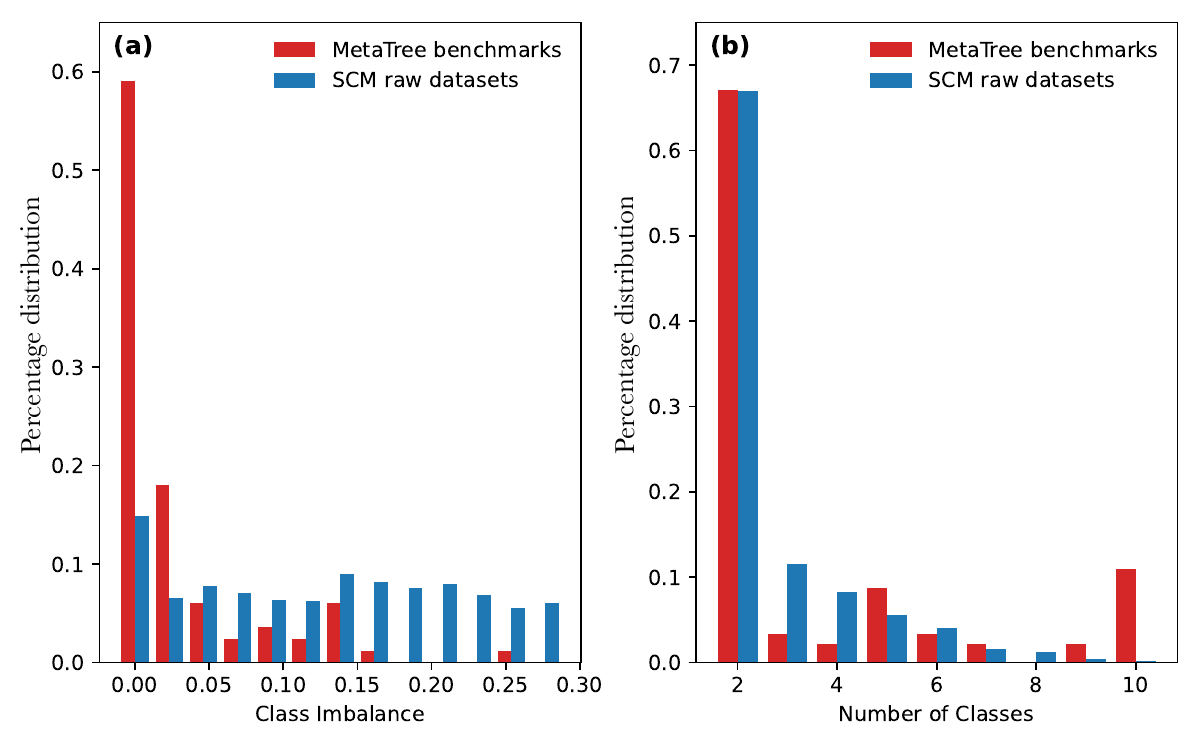}
    \caption{
    Class imbalance distribution comparison between MetaTree benchmarks and the synthetic datasets generated using the proposed workflow. 
    The synthetic dataset display more uniform class distribution between the selected range [0,0.3] compared to hand-curated MetaTree benchmarks.
    Class count distribution comparison between MetaTree benchmarks and the synthetic datasets generated using the proposed workflow.
    Compared to MetaTree benchmarks, which were manually hand-picked and hence display irregular distribution, the class count distribution in synthetic datasets can be well-explained by the quality filters we enforced. Both accuracy and class imbalance filters will favor datasets with smaller number of classes.
    }
    \label{fig:scm_data_quality}
\end{figure}

\subsection{More details on quality filters for improving the pre-training data}
\label{details_qual_filters}

Raw synthetic data from the SCM pipeline is often unsuitable for building decision trees. To address this, we implemented additional data-quality filters. These filters ensure that datasets used to generate pre-training targets for meta-learning are well-suited for decision tree construction, specifically by avoiding class imbalance and enabling the generation of trees that accurately classify ground truth labels.

High class imbalance during pre-training is undesirable because it frequently results in trivial decision stumps, where tree algorithms such as GOSDT or CART merely predict the majority label for the entire dataset. To counteract this, we introduced a class imbalance filter, ensuring that the majority class comprises no more than 75\% of the samples. This approach enables efficient synthetic data generation while preventing trivial problems during pre-training. Our class imbalance metric is adopted from the Penn Machine Learning Benchmarks (PMLB) [\cite{olson2017}], which is defined as

\begin{equation}
	I = K \sum_{i=1}^K \left( \frac{n_i}{N} - \frac{1}{K} \right)^2
	\label{eq:class_imbalance}
\end{equation}

where, $K$ represents the total number of classes. The variable $n_i$ signifies the number of samples within class $i$, and $N$ denotes the total number of samples present in the dataset.
To ensure the class imbalance value remains within the range of 0 to 1, the value of I in Eq. \ref{eq:class_imbalance} was normalized.
This normalization was performed using a worst-case scenario where all samples share the same class label.
Formally,

\begin{equation}
	I_\mathrm{normalized} = \frac{ I_\mathrm{raw} }{ I_\mathrm{worst\ case} }
	\label{eq:class_imbalance_normalized}
\end{equation}

where, $I_\mathrm{raw}$ is the raw pmlb class imbalance metric from Eq. \ref{eq:class_imbalance}, $I_\mathrm{worst\ case}$ is the worst case class imbalance value if all the samples were to be in just one class($I_\mathrm{worst\ case}=k-1$), and $I_\mathrm{normalized}$ is the normalized class imbalance metric.
To ensure no majority class contains more than 75\% of the samples, we filtered for SCM dataset with $I_\mathrm{normalized}$ value less than 0.3 for generating the pre-training target.

Similarly, to ensure the suitability of candidate SCM datasets for decision tree generation, we implemented an accuracy filter.
We fitted CART trees to the raw SCM datasets, retained only those where the CART trees achieved over 70\% accuracy.
This, in conjunction with the previously introduced class imbalance filter, enables us to select datasets that facilitate the generation of relatively high-accuracy decision trees, while simultaneously preventing oversampling of datasets with significant class imbalance.

\subsection{Supporting information for diversity of SCM data}
\label{scm_diversity}

The proposed synthetic data generation pipeline provides enhanced control over dataset diversity and quality, allowing for the creation of pretraining datasets with more diverse class imbalances and systematic label distributions.
As illustrated in Fig. \ref{fig:scm_data_quality} (see Appendix), our synthetic datasets exhibit a more uniform class imbalance distribution across the [0, 0.3] range compared to the hand-curated MetaTree benchmark.
Furthermore, the class count distribution shows a systematic decline as the number of classes increases.
This trend is an intended consequence of our accuracy and class imbalance quality filters, which inherently favor datasets with fewer classes.
In essence, our pipeline offers better control over data diversity and quality compared to the manual curation of public datasets.

\section{Acknowledgements}
We would like to thank our colleague, Doron Bergman, for his careful review of the manuscript and his insightful suggestions.

\newpage


\end{document}